\documentclass[11pt]{article}

\usepackage[preprint]{acl}

\usepackage{times}
\usepackage{latexsym}

\usepackage[T1]{fontenc}

\usepackage[utf8]{inputenc}

\usepackage{microtype}

\usepackage{inconsolata}

\usepackage{graphicx}
\usepackage{longtable}
\usepackage{amsmath}
\usepackage{array}
\usepackage{tabularx}
\usepackage{booktabs}
\usepackage{adjustbox}
\usepackage[utf8]{inputenc}
\usepackage{amsfonts}
\usepackage{pgfplots}
\usepackage{natbib}
\usepackage{tcolorbox}
\tcbuselibrary{breakable}
\pgfplotsset{compat=1.18}

\title{SRR-Judge: Step-Level Rating and Refinement for Enhancing Search-Integrated Reasoning in Search Agents}

\author{Chen Zhang, Kuicai Dong, Dexun Li,  \\ \textbf{Wenjun Li},
\textbf{Qu Yang}, \textbf{Wei Han}, \textbf{Yong Liu} \\
  Huawei Technologies \\
  \texttt{zhang.chen4@huawei.com}
}

\begin{document}
\maketitle
\begin{abstract}

Recent deep search agents built on large reasoning models (LRMs) excel at complex question answering by iteratively planning, acting, and gathering evidence, a capability known as search-integrated reasoning. However, mainstream approaches train this ability often using only outcome-based supervision, neglecting the quality of intermediate thoughts and actions. We introduce SRR-Judge for reliable step-level assessment of reasoning and search actions. Integrated into a modified ReAct-style rate-and-refine workflow, SRR-Judge provides fine-grained guidance to search-integrated reasoning and enables efficient post-training annotation. Using SRR-annotated data, we apply an iterative rejection-sampling fine-tuning procedure to enhance the deep-search capability of the base agent. Empirically, SRR-Judge delivers more reliable step-level evaluations than much larger models such as DeepSeek-V3.1, with its ratings showing strong correlation with final-answer correctness. Moreover, aligning the policy with SRR-Judge–annotated trajectories leads to substantial performance gains, yielding over a 10\% average absolute pass@1 improvement across the challenging deep-search benchmarks.

\end{abstract}

\section{Introduction}


Large language models (LLMs) have rapidly evolved from one-shot text generators into agentic foundation models that actively plan and act~\citep{deepseek-ai2025deepseek0v3020,team2025glm04050}. In the new paradigm, an LLM alternates between internal reasoning and external tool executions such as web searches, API calls, and code execution, in a loop~\citep{yao2023react}. In information-seeking scenarios, this approach excels at uncovering hard-to-find information by utilizing search tools and allowing the LLM to focus on autonomous reasoning, reflection, cross-validation, and dynamic tool-use decisions, referred as search-integrated reasoning~\citep{li-etal-2025-search,jin2025searchr1trainingllmsreason}. This direction has sparkled a series of works on deep search agents~\citep[inter alia]{li2025webthinkerempoweringlargereasoning,wu2025webdancer,sun-etal-2025-simpledeepsearcher,zheng-etal-2025-deepresearcher,tongyideepresearchteam2025tongyideepresearchtechnicalreport,2025mirothinker,shi2025deepdiver0,lu2025deepdiveadvancingdeepsearch}. 


Majority of the works rely on the end-to-end reinforcement learning using verifiable rewards (RLVR)~\citep{shao2024deepseekmath} to optimize the models' search-integrated reasoning capability. However, this form of sparse, outcome-based supervision provides low reward density, neglecting the coherence, conciseness, and logical structure of intermediate reasoning steps and providing no correction on problematic intermediate search actions~\citep{zhang2025process}. Process-level supervision offers a natural remedy and has shown strong promise across several domains, including math reasoning~\citep{lightman2024lets,wang-etal-2024-math,zhang-etal-2025-lessons}, code generation~\citep{li-etal-2025-codeprm,yu2025reasoning}, and simulated long-horizon agentic tasks~\citep{xiong-etal-2024-watch,xi2025agentprm0,miao2025boosting} in a controlled environment, such as web shopping and TextWorld. In these domains, intermediate steps can be easily verified and the action space is relatively constrained to allow reliable step-level evaluation. In contrast, deep-search agents operating in real web environments face a fundamentally different setting: the interaction environment is effectively a black box, and there is no notion of ground-truth search actions. Even small perturbations in search queries can lead to discontinuous changes in retrieved content~\citep{ning2025survey}. This fragility makes it difficult to define consistent step-level signals and limits the applicability of existing process-level supervision frameworks beyond controlled or simulatable domains.

In this paper, we address this gap by developing process-level supervision tailored to complex deep-search scenarios. We introduce SRR-Judge, an automatic generative evaluator that provides fine-grained feedback on both step-level reasoning and search actions, scoring intermediate steps and triggering targeted refinements when low-quality behavior is detected. By identifying faulty actions throughout long-horizon trajectories, SRR-Judge helps steer the iterative search process toward more coherent and effective reasoning. The key idea of SRR-Judge is to distill step-level evaluation capabilities from a much larger agentic foundation model into a smaller, more efficient reasoning model. These agentic foundation models already encode rich knowledge of search behavior and reasoning quality from training on extensive agentic trajectories, tool-use logs, and web interactions, making them natural teachers for step-level judgment. Furthermore, distillation enables cost-effective, fine-grained supervision at scale during training trajectory annotation and deployment, without requiring expensive human annotations.

Using SRR-Judge, we adapt the ReAct workflow~\citep{yao2023react} into a step-level rate-and-refine inference workflow (\S\ref{subsec:SRR-workflow}) for producing higher-quality search-integrated reasoning trajectories. These trajectories are subsequently used to optimize the agent’s deep search capability through an iterative rejection-sampling fine-tuning procedure (RFT) (\S\ref{subsec:sa-alignment}). In our experiments, we show that SRR-Judge achieves evaluation performance comparable to that of the much larger teacher model, as demonstrated by strong point-biserial correlations between step-level ratings and final trajectory success. Moreover, experimental results indicate that the rate-and-refine workflow substantially improves the deep-search capabilities of weak search agents at inference time. Finally, applying the same workflow to generate training trajectories for iterative alignment yields an absolute average pass@1 gain of more than 10\% for the base policy on complex real-world information-seeking benchmarks.

In summary, we make the following contributions: (1) We propose SRR-Judge for step-level evaluation of search-integrated reasoning along a long horizon agent trajectory; (2) We carefully analyze the relationship between SRR-Judge's step-level evaluation and the trajectory-level outcome, showing that the step-level ratings are strong indicators of outcome correctness. (3) Through SRR-Judge, we propose a rate-and-refine workflow for trajectory data quality improvement, which significantly boots the performance of search agent through an iterative rejection sampling procedure.

\section{Methodology}

In \S\ref{subsec:infer-workflow} and \S\ref{subsec:srr-judge}, we introduce the two preliminary components of our proposed rate-and-refine inference workflow (\S\ref{subsec:SRR-workflow}). In \S\ref{subsec:sa-alignment}, we describe how SRR-judge helps enhance the search-integrated reasoning of agentic search models via iterative rejection-sampling fine-tuning. 

\subsection{Basic Inference Workflow}
\label{subsec:infer-workflow}

Following~\citep{li-etal-2025-search}, we adopt the ReAct-based~\citep{yao2023react} inference workflow where an LLM-empowered search agent $\mathcal{M}$ engages in multi-turn interactions with the search environment, which comprises search-related tools $\mathcal{T}$. Specifically, (1) a Google Serper API ($\mathcal{T}_s$) that returns Top-10 relevant urls and their corresponding snippets for a specific search query, and (2) an open-source web-scraping tool for content extraction from the webpages. 

Given a question $q$, $\mathcal{M}$ autonomously reasons and initiates actions, either invoke the search tools ($act^s$) or provide the answer if existing information is deemed sufficient ($act^a$). For each search action \( act^s \), the search environment returns the corresponding observation \( o \), which is obtained by providing the search query and scraped raw webpage as inputs to a large language model (LLM) that summarizes the key information relevant to the query. The entire reasoning trajectory $\mathcal{I}_q$ consists of $\left\{(t_i, act^s_i, o_i)_{i=0}^{k}, \, t^a, \, act^a \right\}$, where the tuple $(t_i, act^s_i, o_i)$ denotes the interleaved triples of thought, action, and observation, iterating from $i = 0$\footnote{No invocation of search tools} to $k$ steps until the final reasoning and answer $(t^a, act^a)$ are reached or $k$ exceeds $K$\footnote{Due to cost consideration, $K$ is defined as 20 throughout the paper unless otherwise specified.}, a predefined maximum number of interactions.

\subsection{SRR-Judge}
\label{subsec:srr-judge}

\paragraph{What SRR-Judge Evaluate?}
For simplicity, we use \( t_j \) and \( act_j \) to represent the step-level reasoning and the corresponding action at step \( j \), where \( act_j \) can be either the search or the answer action. SRR-Judge is a generative judge designed to evaluate and refine the step-level reasoning within \( \mathcal{I}_q \), as defined by the function:

\[
(e_j, r_j, \tilde{t}_j, \tilde{act}_j) = F(h_j, t_j, act_j)
\]
Specifically, it outputs the judge explanation $e_j$ and quality rating \( r_j \) for \( t_j \) and \( act_j \), conditioned on the history reasoning trajectory \( h_j = (t_i, act_i, o_i)_{i=0}^{j-1} \). Meanwhile, $t_j$, $act_j$ are refined as $\tilde{t}_j$ and $\tilde{act}_j$ accordingly. The judge prompt template is outlined in Table~\ref{tab:SRR-judge-instruction-prompt} of the Appendix~\ref{sec:prompt-template}. Specifically SRR-Judge evaluates each step's reasoning and action using four criteria: (1) \textbf{Clarity \& Conciseness}: Whether the reasoning is direct, easy to follow, and free of redundancy; (2) \textbf{Logical Structure}: Whether the step is coherent with the history trajectory and advances the solution to the user's query; (3) \textbf{Query Appropriateness}: If \( act^s \) is invoked, whether the search query reflects the intended search goal and uses appropriate keywords. If \( act^a \) is invoked, whether the final answer is faithful to information retrieved in earlier steps; (4) \textbf{Coverage \& Improvement Potential}: Whether the step addresses key aspects of the user's question and whether further improvement (e.g., additional sources or a refined search strategy) is possible.

\paragraph{Judgment Data Annotation}
A central challenge for training automatic judges is the scarcity of annotated data. To address this, we adopt a straightforward strategy: leveraging a strong agentic foundation model, DeepSeek-V3.1, combined with the self-consistency technique~\citep{wang2023selfconsistency}, to annotate step-level reasoning. These annotations are generated from reasoning trajectories produced by running different models through the inference workflow described in~\ref{subsec:infer-workflow}, using the prompt template provided in Table~\ref{tab:SRR-judge-instruction-prompt}. Collecting reasoning trajectories from different models helps increase the diversity of step-level judge data for annotation. The motivation for using strong agentic foundation models for annotation is that they are capable search agents with the knowledge needed to judge whether search actions are appropriate, while the self-consistency technique helps mitigate single-run evaluation bias. The step-level final rating $r_j$ is determined by a majority vote across five independent annotation runs of DeepSeek-V3.1 over the reasoning trajectories. Other annotations, including \( e_j \), \( \tilde{t}_j \), and \( \tilde{act}_j \), are randomly selected from one of the majority-vote runs. Further implementation details on data annotations are described in \S\ref{subsec:training-data}.

\paragraph{Judgment Data Quality Enhancement}
Purely relying on model annotation and self-consistency does not fully guarantee the quality of the annotated data. Yet, the cognitive load required for humans to perform large-scale quality checks on these annotations is substantial. To address this issue, we resort to a ``silver" approach by using point-biserial correlation coefficients to measure the relationship between the annotations and the verifiable binary correctness of the entire reasoning trajectories. The step-level ratings along a search-integrated reasoning trajectory are aggregated by averaging. Intuitively, trajectories with lower average ratings are expected to yield incorrect predictions, whereas higher averages indicate correct predictions. We use a point-biserial threshold of 0.7 to balance data quantity and quality, filtering out trajectories whose average ratings conflict with their final correctness. 
\paragraph{Training SRR-Judge}
With the step-level annotated data, we fine-tune QwQ-32B for one epoch using the standard SFT loss to produce SRR-Judge. Two major considerations behind of not directly using the DeepSeek-V3.1 as the judge. First, DeepSeek-V3.1 is substantially more computationally expensive, making it impractical for frequent step-level evaluation in our rate-and-refine inference workflow (\S\ref{subsec:SRR-workflow}). Second, our empirical results (\S\ref{subsec:ds-qa}) indicate that the fine-tuned SRR-Judge is sufficient to achieve comparable performance to using DeepSeek-V3.1 as the judge. We select a 32B model over smaller models, such as Qwen3-8B, is due to its stronger reasoning and better text generation/instruction-following capabilties.

\subsection{Step-Level Rate-and-Refine Inference Workflow}
\label{subsec:SRR-workflow}

\begin{figure}[t]
    \centering
    \includegraphics[width=\linewidth]{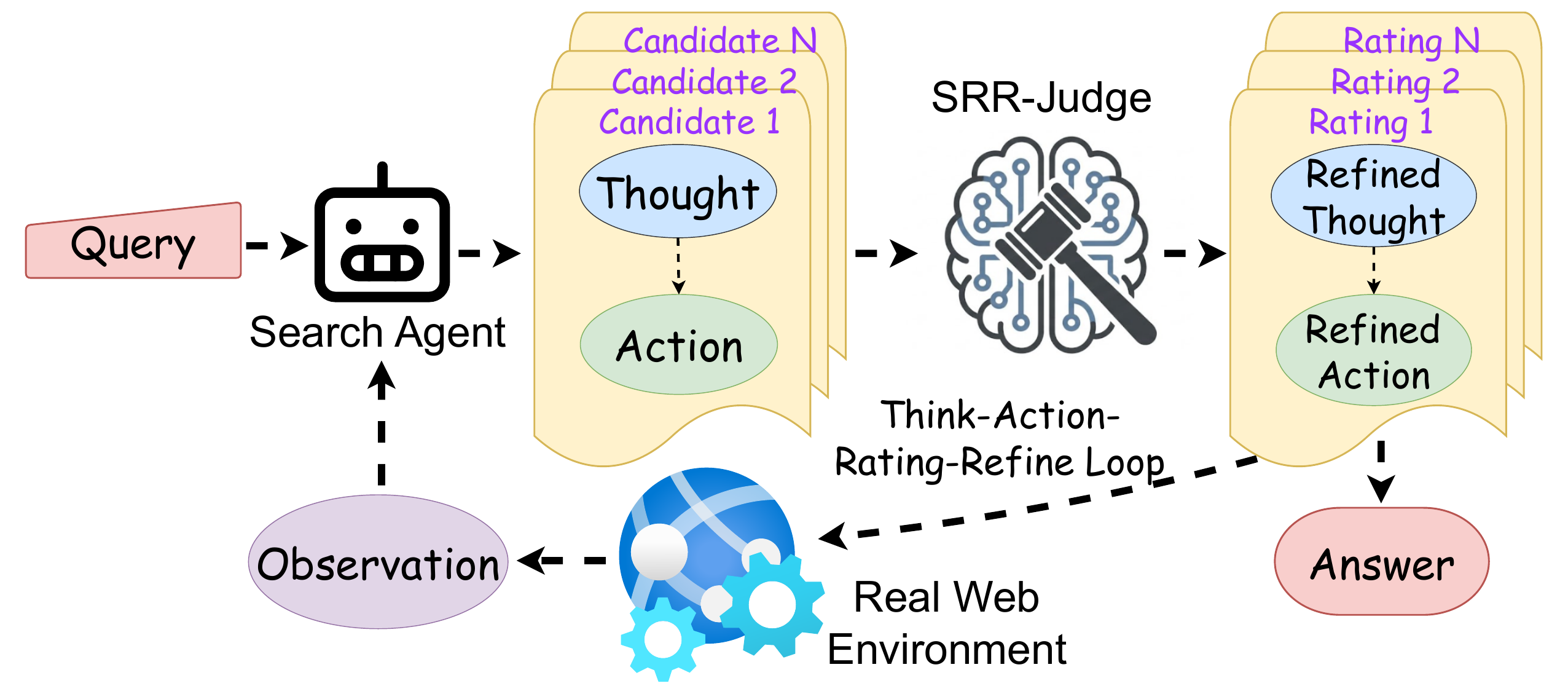}
    \caption{The rate-and-refine inference workflow.}
    \label{fig:srr-inference}
\end{figure}

Figure~\ref{fig:srr-inference} illustrates the step-level rate-and-refine inference workflow, a direct modification of ReAct with SRR-Judge. This workflow enables step-level best-of-N sampling and thought–action refinement. Specifically, SRR-Judge assigns ratings to each step-level thought–action candidate, and the inference process continues with the highest-scored candidate. If none of the candidate receives a rating above the threshold rating of 4, SRR judge refines and improves the highest-scored candidate. The iterative process continues when an answer is generated or the max loop count is reached. For fast online inference, we set $N = 1$, whereas for generating high-quality trajectories via Best-of-$N$ sampling to support search-agent alignment (\S\ref{subsec:sa-alignment}), we adopt a larger value of $N = 5$. 

\subsection{Search Agent Alignment With SRR-Judge}
\label{subsec:sa-alignment}


Given the high cost of mainstream agentic online reinforcement-learning methods, such as GRPO~\citep{shao2024deepseekmath,jin2025searchr1trainingllmsreason} and PPO~\citep{schulman2017proximal}, that require continuous interaction with real web environments through API calls or high-fidelity simulated environments during training, we adopt an offline alignment strategy in which the base model is first trained with SFT and then improved through an iterative RFT procedure. In each iteration, the current policy generates new trajectories that are annotated and refined by SRR-Judge through the rate-and-refine workflow. The resulting high-quality data are then combined with trajectories from previous iterations and used for the next round of RFT, allowing the policy to progressively improve over successive iterations. 

\section{Experimental Setup}

\subsection{Training}
\label{subsec:training-data}

\begin{table}[t]
\centering
\resizebox{\linewidth}{!}{
\begin{tabular}{lccc}
\toprule
\textbf{Dataset} & \textbf{Full} & \textbf{Sampled} & \textbf{Purpose} \\
\midrule
    WebDancer~\citep{wu2025webdancer} & 200 & 200 & Judge/SFT \\
    WebShaper~\citep{tao2025webshaper0} & 500 & 500  & Judge/SFT \\
    WebExploxer~\citep{liu2025webexplorerexploreevolvetraining} & 100 & 100  & Judge/SFT \\
    DeepDive~\citep{lu2025deepdiveadvancingdeepsearch} & 3,250 & 3,250 & Judge/SFT \\
    InfoSeek-Hard~\citep{xia2025opendatasynthesisdeep} & 18,476 & 5,000 & Judge/SFT/RFT \\
    DuetQA-Verified~\citep{singh2025fathomdeepresearchunlockinglonghorizon} & 4,889 & 4,889 & Judge/SFT/RFT \\ 
    ASearcher-LRM~\citep{gao2025turnsunlockinglonghorizonagentic} & 35,054 & 6,000 & SFT/RFT \\ 
    \midrule
    BrowseComp-En~\citep{wei2025browsecompsimplechallengingbenchmark} & 1,266 & 1,266 & Evaluation \\
    BrowseComp-ZH~\citep{zhou2025browsecompzhbenchmarkingwebbrowsing} & 289 & 289 & Evaluation \\
    Xbench-DeepSearch~\citep{chen2025xbenchtrackingagentsproductivity} & 100 & 100 & Evaluation \\
InfoDeepSeek-ZH~\citep{xi2025infodeepseekbenchmarkingagenticinformation} & 245 & 245 & Evaluation \\
\bottomrule
\end{tabular}
}
\caption{Dataset statistics, including the full size of each dataset (\# of QA pairs), the subsampled portion used for training/evaluation, and their intended use.}
\label{tab:train-data-stats}
\end{table}

The training data of both SRR judge and search agent alignment are derived from a diverse mixture of open-source complex question-answering (QA) datasets. Details are outlined in Table~\ref{tab:train-data-stats}. 

\begin{figure}[!t]
    \centering
    \includegraphics[width=\linewidth]{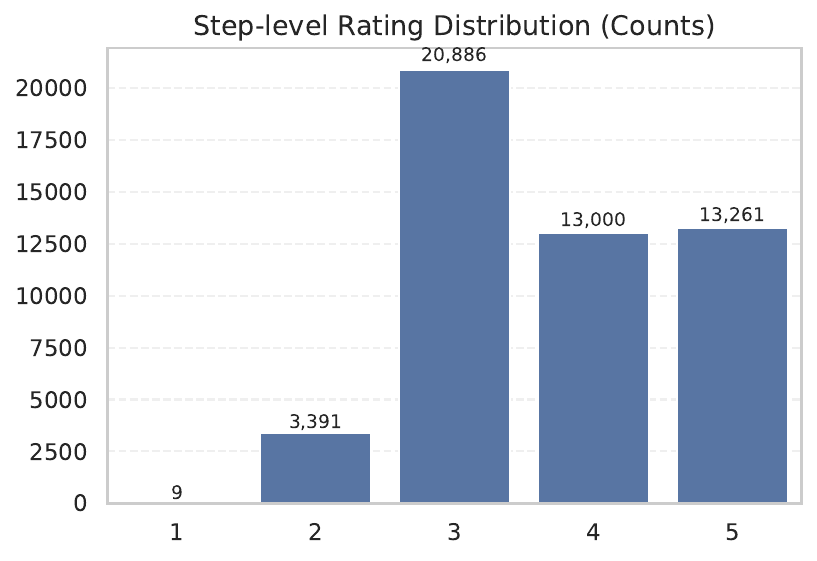}
    \caption{Distribution of step-level ratings.}
    \label{fig:rating-dist}
\end{figure}

\paragraph{SRR Judge Training Data} To train SRR-Judge, we sample 6,050 QA pairs from the dataset mixture and run four model variants (DeepSeek-V3.1, Qwen3-235B-A22B, QwQ-32B, and Qwen3-8B) through the ReAct workflow to obtain their full search-integrated reasoning trajectories. The average number of interactions per trajectory\footnote{average number of step-level data points per trajectory for annotation.} between the search agent and the web search API is 4.37 across all model variants. All trajectories are evaluated for binary answer correctness using LLM-as-a-judge. Step-level data annotation and filtering are performed based on the procedure described in \S\ref{subsec:srr-judge}. As shown in Figure~\ref{fig:rating-dist}, rating 3 dominates the dataset, while rating 1 is extremely scarce (only 9 samples). To counter class imbalance, we subsample 10{,}000 items from ratings 3–5 and upsample rating-2 data by a factor of three. For rating 1, we form negatives by pairing each trajectory history with an unrelated thought–action candidate and annotating it using DeepSeek-V3.1, producing 10{,}000 additional rating-1 instances.

\paragraph{Search Agent Training Data} For SFT training, we reuse the 6,050 DeepSeek-V3.1 trajectories described above and supplement them with an additional 2,000 DeepSeek-V3.1 trajectories generated using QA pairs from ASearcher-LRM. After filtering out trajectories that produce incorrect final answers or contain malformed formats, a total of 5,039 trajectories are retained. The average number of search API calls per trajectory is 6.46. 

For the iterative RFT procedure, each iteration processes roughly 6K QA pairs using the current policy model under the rate-and-refine workflow. We repeat this process for two iterations. After filtering out incorrect or malformed trajectories, the remaining usable trajectory numbers are 3,581 and 2,650 for iterations one and two, with the corresponding trajectories containing an average of 5.93 and 6.05 API calls, respectively.

\begin{figure*}[t]
    \centering
    \includegraphics[width=0.8\linewidth]{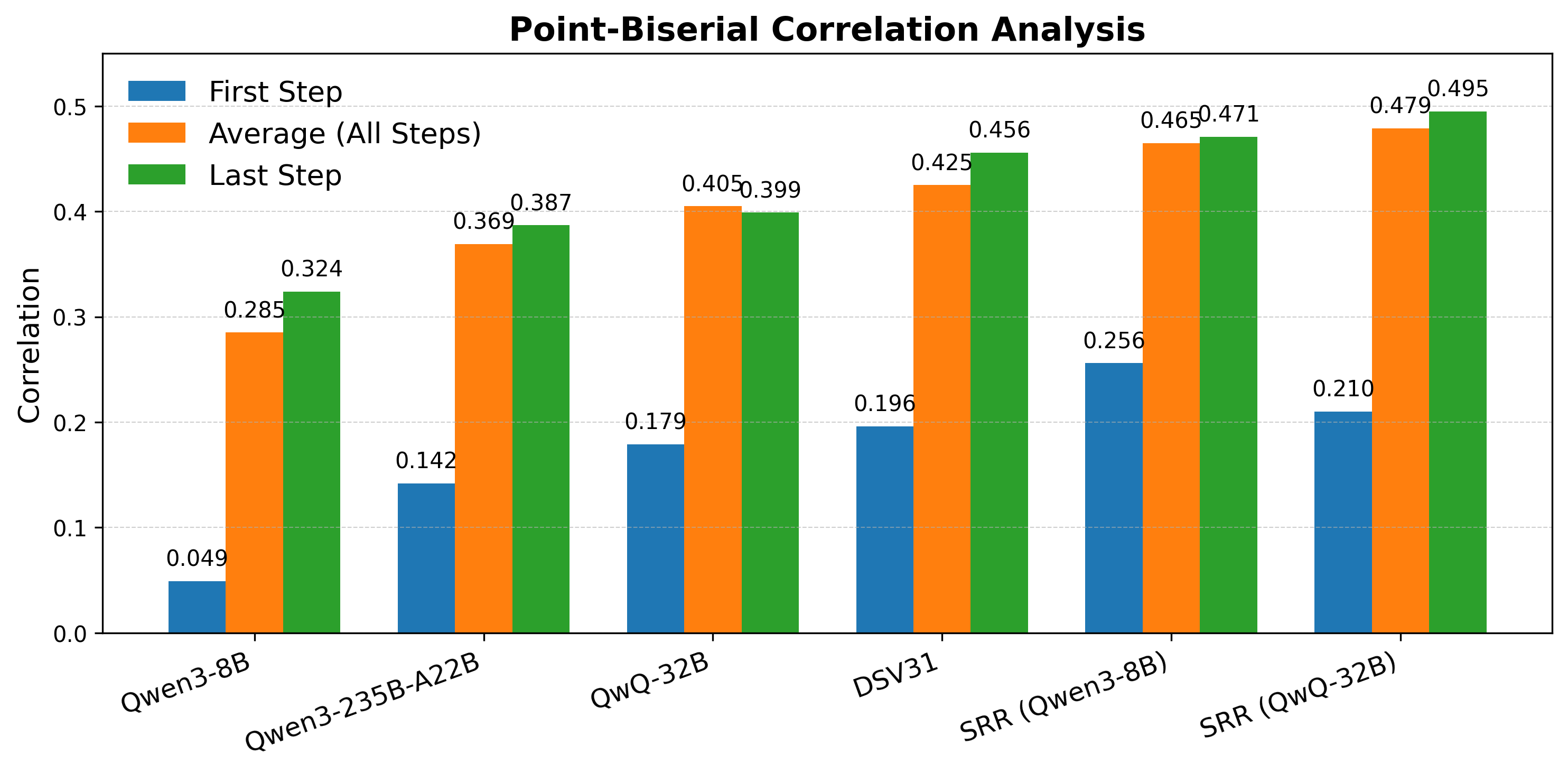}
    \caption{Correlation performance of the judge models on first-round, last-round, and averaged step-level ratings.}
    \label{fig:correlation-analysis}
\end{figure*}

\subsection{Evaluation}
\label{subsec:eval-data}

SRR-Judge is evaluated by correlating its step-level ratings with the verifiable trajectory-level answer correctness. Using QA pairs from the 4 evaluation datasets listed in Table~\ref{tab:train-data-stats}, we run 4 model variants (same as the training data construction) through the ReAct workflow, yielding 3,736 search-integrated trajectories with an average of 6.12 search API calls per trajectory. We then examine the correlations between trajectory-level correctness and three rating statistics from SRR-Judge: the first-step rating, the last-step rating, and the average rating across all steps. For search agent alignment, we assess their end-to-end performance on three real-web information-seeking benchmarks, BrowseComp, BrowseComp-Zh, and XBench-DeepSearch, by measuring pass@1 accuracy.

\section{Results \& Analysis}
\label{sec:results}

\subsection{Correlation Analysis}
\label{subsec:correlation-analysis}

In the correlation analysis, SRR-Judge fine-tuned from QwQ-32B and Qwen3-8B are compared to other LLM judges, including Qwen3-8B, QwQ-32B, Qwen3-235B-A22B, and DeepSeek-V3.1. Figure~\ref{fig:correlation-analysis} depicts their respective first-step, last-step, and average-across-all-steps point-biserial correlation scores. Both SRR variants achieve substantially higher first-step, last-step, and average point-biserial correlation scores than their respective base models, indicating that SRR fine-tuning greatly enhances a model’s ability to judge step-level search-integrated reasoning. Notably, SRR (QwQ-32B) achieves the highest overall performance, with the strongest average (0.479) and last-step correlation (0.495) among all judges evaluated. This suggests that QwQ-style slow-thinking reasoning models provide a particularly effective foundation for judge model training. Furthermore, the SRR judges outperform much larger general-purpose models, including Qwen3-235B-A22B and even DeepSeek-V3.1, the source of distilled supervision, demonstrating the effectiveness of our quality-enhancement procedure and the importance of task-specific fine-tuning for search-integrated reasoning.

A notable pattern across all LLM judges is that correlation scores significantly increase from the first step to the last step, indicating that judge ratings of later steps provide a clearer and more reliable signal of task success. At the same time, correlations of the average-across-steps remain close to the final-step scores, suggesting that intermediate judgments contribute meaningful reward signals and further motivate the utility of our proposed step-level rate-and-refine inference workflow.

It is worth noting that all LLM judges remain below 0.5 correlation because final-answer correctness is shaped by factors beyond search-integrated reasoning alone, including environment returns, stochastic exploration behavior, and other non-reasoning components of the task pipeline that lie outside the SRR judge’s evaluative scope. Consequently, judge–answer correlation is inherently upper-bounded, even for strong evaluators.

\subsection{Qualitative Analysis}
\label{subsec:human-eval}

\begin{figure}[!t]
\small
\fbox{
\begin{minipage}{0.96\linewidth}

\textbf{Question.} \\
The artist’s father was a Protestant minister. They had a brief marriage to a colloquial poet who studied engineering and was born in September between 1938--1941. Later, they married a fashion designer. In the 1980s, they established a school for the art they were known for in a village where that art had not previously existed. What is the artist’s name?

\medskip
\textbf{Answer.} \\
Evelyne Porret

\medskip
\textbf{Original Reasoning and Search Action (Step 1).} \\
``Okay, let's try to figure out this question step by step. [...] The artist had a brief marriage to a colloquial poet who studied engineering and was born in September between 1938--1941. [...]

\textcolor{red}{Let me start by thinking about famous artists from that era.} [...]
\textcolor{red}{Allen Ginsberg?} [...]
\textcolor{red}{John Lennon?} [...]
\textcolor{red}{Miroslav Holub?} [...]
\textcolor{red}{Richard Brautigan?} [...]
\textcolor{red}{Ted Hughes?} [...]
\textcolor{red}{Sylvia Plath?} [...]

\textcolor{red}{Another angle: the artist married a fashion designer.} [...] 
\textcolor{red}{Yoko Ono?} [...]

\textcolor{red}{Maybe the artist started a school in the 1980s.} [...] 
\textcolor{red}{Perhaps a pottery village, or a specific craft.} [...]

\textcolor{red}{Let me try to tackle the spouses again.} [...]

\textcolor{red}{Let's do a search for poets born in September between 1938--1941.}'' \\
\textit{Search Query:} \texttt{poets born September 1938 to 1941}

\medskip
\textbf{SRR-Judge Evaluation.} \\
\textit{Rating: 3/5.} \\
``The reasoning trace is somewhat clear but \textcolor{blue}{lacks conciseness due to listing multiple artists without strong relevance}. The logical structure is weak as it \textcolor{blue}{jumps between clues without prioritizing unique identifiers}. The search query is appropriate for the date range but \textcolor{blue}{omits key qualifiers such as engineering background or colloquial style}. Improvement is needed by \textcolor{blue}{focusing on the art school clue} and connecting it to the artist’s identity.''

\medskip
\textbf{SRR-Judge Refined Thought and Action.} \\
``\textcolor{blue}{The poet’s engineering background and colloquial style are distinctive. I should identify this poet first, then find artists briefly married to them. The father being a Protestant minister, the later marriage to a fashion designer, and the art school founded in the 1980s can be used as filters.}'' \\
\textit{Search Query:} \texttt{colloquial poet born September 1938 engineering degree}

\end{minipage}
}
\caption{Artist example from BrowseComp. Step~1 of a search-integrated reasoning trajectory generated by DeepSeek-R1, evaluated and refined by SRR-Judge.}
\label{fig:appendix-srr-artist-step1}
\end{figure}

Through manual inspection of SRR-enabled rate-and-refine trajectories, we find that SRR-Judge encourages agents to prioritize highly discriminative constraints early, rather than engaging in speculative candidate enumeration. Furthermore, when searches stall in repetitive searches, the judge provides mid-trajectory correction by refocusing the policy on systematic verification of bounded candidate sets. At stages closer to the end of the trajectory, the search agent is guided to synthesize accumulated evidence and make principled decisions grounded on available information that satisfy existing constraints. Due to space constraints and the long-horizon nature of search-integrated trajectories, we show an example of SRR-Judge correcting early-stage deep-search reasoning in Figure~\ref{fig:appendix-srr-artist-step1}, where the highlighted text (red) marks speculative enumeration and weak constraint prioritization in the original reasoning, while the refinement provided by SRR-Judge (blue) reorients the search toward the most discriminative clues. We defer other detailed examples to Appendix~\ref{sec:qa-srr-additional-example}.

\subsection{Deep Search QA}
\label{subsec:ds-qa}

\paragraph{Comparison with the Teacher}

\begin{table}[t]
\centering
\small
\resizebox{\linewidth}{!}{
\begin{tabular}{lcc}
\toprule
& \textbf{SRR-Judge} & \textbf{DeepSeek-V3.1} \\
\midrule
BrowseComp & 9.1 $\pm$ 1.1 & 11.3 $\pm$ 1.3 \\
BrowseComp-ZH & 29.2 $\pm$ 2.3 & 29.1 $\pm$ 2.7 \\
Xbench-DeepSearch & 53.3 $\pm$ 4.2 & 58.0 $\pm$ 4.0 \\
\midrule
\end{tabular}
}
\caption{Performance comparison (average pass@1 \%) of SRR and the teacher executing the rate-and-refine workflow on the QwQ-32B base model.}
\label{tab:refine-comparison}
\end{table}

Table~\ref{tab:refine-comparison} presents the performance differences between using SRR-Judge and DeepSeek-V3.1 (the teacher) in refining the performance of QwQ-32B model. Although DeepSeek-V3.1 achieves slightly higher refinement performance, it is substantially larger in scale, whereas our SRR-Judge delivers comparable gains with a much smaller 32B model, making it substantially more efficient for deployment and more practical for large-scale, low-cost post-training data annotation.

\paragraph{SRR-Judge Improves Search Agents at Inference Time} Table~\ref{tab:main-result} shows that SRR-Judge consistently improves search agents at inference time across all benchmarks. For QwQ-32B, SRR-based refinement raises Pass@1 from 5.4 to 9.1 on BrowseComp, 23.4 to 29.2 on BrowseComp-ZH, and 46.0 to 53.3 on Xbench-DeepSearch, without any additional training. Similar gains are observed for DeepSeek-R1 and QwQ-32B-SFT, demonstrating that SRR-Judge generalizes across base models. These improvements are achieved purely through inference-time refinement, showing that SRR-Judge can correct reasoning and search errors on the fly. Despite using the same backbone, the rating \& refinement approach matches stronger RLVR-based open-source deep-search implementations (e.g., WebSailor-32B and DeepDive-32B), highlighting the effectiveness of step-level guidance for enhancing search-integrated reasoning without modifying model parameters.

\begin{table*}[t]
\centering
\resizebox{0.85\linewidth}{!}{
\begin{tabular}{lccc}
\toprule
\textbf{Model} & \textbf{BrowseComp} & \textbf{BrowseComp-ZH} & \textbf{Xbench-DeepSearch} \\
\midrule
\multicolumn{4}{c}{\textbf{\textit{Proprietary Agentic Foundation Models with Tools}}} \\
\midrule
OpenAI-o3  &  49.7  & 58.1 &  67.0  \\
Claude-4.5-Sonnet            & 19.6  & 40.8 & 66.0  \\
Grok-DeepResearch             & -     & 12.9 & 50.0  \\
Doubao-DeepThink              & -     & 26.0 & 50+\\
OpenAI-DeepResearch           & 51.5  & 42.9 & -  \\
\midrule
\multicolumn{4}{c}{\textbf{\textit{Open-Source Agentic Foundation Models with Tools}}} \\
\midrule
GLM-4.6                       & 45.1 & 49.5 & 70.0  \\
DeepSeek-V3.1                 & 30.0 & 49.2 & 71.0 \\
Kimi-K2-0905                  & 7.4  & 22.2 & 61.0\\
Tongyi-DR-30B~\citep{tongyideepresearchteam2025tongyideepresearchtechnicalreport}       &  43.4 & 46.7 & 75.0 \\
MiroThinker-v1.0-30B~\citep{2025mirothinker}          & 41.2 & 47.8 & 70.6 \\
\midrule
\multicolumn{4}{c}{\textbf{\textit{Open-Source Deep Search Solutions}}} \\
\midrule
Search-O1-32B~\citep{li-etal-2025-search}                 & 2.8  & 17.9 & 25.0 \\
ASearcher-Web-32B~\citep{gao2025turnsunlockinglonghorizonagentic}             & 5.2  & 15.6 & 42.1 \\
WebThinker-32B~\citep{li2025webthinkerempoweringlargereasoning}                & 2.8  & 7.3  & 24.0  \\
AFM-32B~\citep{li2025chain0of0agents0} & 11.1 & - & -  \\
WebDancer-32B~\citep{wu2025webdancer}  & 3.8  & 18.0 & 39.0 \\
WebSailor-32B~\citep{li2025websailor0} & 10.5 & 25.5 & 53.3\\
DeepDive-32B~\citep{lu2025deepdiveadvancingdeepsearch} & 15.3 & 29.7 & 51.8 \\ 
\midrule
\multicolumn{4}{c}{\textbf{\textit{Our Implementations}}} \\ \midrule
QwQ-32B & 5.4 $\pm$ 0.5 & 23.4 $\pm$ 2.0 & 46.0 $\pm$ 4.0 \\
QwQ-32B + SRR Refine &  9.1 $\pm$ 1.1 & 29.2 $\pm$ 2.3 & 53.3 $\pm$ 4.2 \\ 
DeepSeek-R1 & 7.3 $\pm$ 0.5 & 36.5 $\pm$ 1.3 & 53.7 $\pm$ 2.3 \\
DeepSeek-R1 + SRR Refine & 14.6 $\pm$ 3.2 & 37.8  $\pm$ 0.7 & 55.3 $\pm$ 2.5 \\ 
QwQ-32B-SFT & 12.6 $\pm$ 0.2 & 31.8 $\pm$ 1.5 & 52.7 $\pm$ 3.0 \\
QwQ-32B-SFT + SRR Refine & 14.4 $\pm$ 0.2  & 32.0 $\pm$ 1.1 & 55.0 $\pm$ 3.5 \\ \midrule
QwQ-32B-RFT-Iter1 (ORM) &  13.9 $\pm$ 0.2 & 32.8 $\pm$ 1.8 & 56.5 $\pm$ 2.1 \\
QwQ-32B-RFT-Iter1 (SRR-Judge) & 15.1 $\pm$ 2.0  & 36.5 $\pm$ 2.7  & 59.7 $\pm$ 1.5 \\
QwQ-32B-RFT-Iter2 (SRR-Judge) & 16.2 $\pm$ 0.8 & 38.3 $\pm$ 2.4 & 61.3 $\pm$ 1.5 \\
\bottomrule
\end{tabular}}
\caption{
Evaluation results - Pass@1 (\%) - on mainstream deep-search QA benchmarks. Results except for our implementations are taken from related works, such as~\citet{lu2025deepdiveadvancingdeepsearch} and~\citet{2025mirothinker}.
For open-source deep research systems, we primarily report 32B-scale models that use a backbone comparable to ours. For our implementations, we report the average Pass@1 and standard deviation across three independent runs. ORM denotes that the training trajectories used for RFT are derived by outcome supervision only.}
\label{tab:main-result}
\end{table*}

\paragraph{SRR-Judge Enables More Effective Alignment Than Outcome-Only Supervision} Comparing QwQ-32B-RFT-Iter1 trained with trajectories selected solely by final-answer correctness (ORM) to those annotated using the SRR rate-and-refine workflow highlights the benefit of step-level supervision. As shown in Table~\ref{tab:main-result}, although both models are fine-tuned from QwQ-32B-SFT using training data of comparable scale, QwQ-32B-RFT-Iter1 (SRR-Judge) achieves higher Pass@1 across all reported benchmarks, improving from 13.9 to 15.1 on BrowseComp, from 32.8 to 36.5 on BrowseComp-ZH, and from 56.5 to 59.7 on Xbench-DeepSearch. These gains indicate that annotating and filtering trajectories with SRR-Judge provides a richer and more informative training signal than trajectory-level correctness alone. By explicitly evaluating and refining intermediate reasoning and search actions, SRR-Judge enables more effective policy alignment for long-horizon deep-search tasks.

\paragraph{Iterative RFT Improves Search Integrated Reasoning.} Table~\ref{tab:main-result} also highlights the effectiveness of iterative RFT with SRR-Judge supervision. Starting from the same QwQ-32B backbone, successive rounds of SRR-guided RFT lead to consistent and monotonic improvements across all benchmarks. Specifically, QwQ-32B-RFT-Iter1 (SRR-Judge) already surpasses QwQ-32B-SFT and the outcome-only RFT. QwQ-32B-RFT-Iter2 further improves performance to 16.2 / 38.3 / 61.3 Pass@1 on BrowseComp, BrowseComp-ZH, and Xbench-DeepSearch, respectively. The results suggest that iterative RFT does not saturate after a single refinement step; instead, SRR-annotated trajectories continue to provide meaningful training signal, supporting scalable improvement in long-horizon search-integrated reasoning.

\section{Related Works}

\subsection{Deep Search Agents}

Works on deep search agents spans several complementary axes, including workflow design~\citep{li-etal-2025-search,qin2025flash0searcher0,li2025webthinkerempoweringlargereasoning,jin2025decoupled}, trajectory data synthesis~\citep{tao2025webshaper0,li2025websailor0,xia2025open,liu2025webexplorerexploreevolvetraining,lu2025deepdiveadvancingdeepsearch}, and advances in training methodologies~\citep{wu2025webdancer,jin2025searchr1trainingllmsreason,zheng-etal-2025-deepresearcher,song2025r10searcher0,shi2025deepdiver0,gao2025turnsunlockinglonghorizonagentic}. Additional efforts focus on building simulated search environments~\citep{sun2025zerosearch0,mei2025o00200searcher0} and developing complex information-seeking benchmarks~\citep{wei2025browsecompsimplechallengingbenchmark,zhou2025browsecompzhbenchmarkingwebbrowsing,chen2025browsecomp0plus0,chen2025xbenchtrackingagentsproductivity,wu-etal-2025-webwalker,pham2025sealqa0}. Beyond academic progress, the field has also gained substantial traction in industry, with notable open-source contributions from teams such as~\citet{tongyideepresearchteam2025tongyideepresearchtechnicalreport} and~\citet{2025mirothinker}. In contrast to most existing methods, which optimize deep search agents through trajectory-level outcome supervision following the RLVR formulation, our approach supplements this coarse feedback with step-level guidance from an automatic generative judge. The judge evaluates each thought–action pair and offers corrections whenever a step is judged to be low quality. 

Several recent studies explore process-level supervision for deep search agents. ReasonRAG~\citep{zhang2025process} introduces process rewards for agentic RAG, using MCTS-based shortest-path estimation to evaluate intermediate search and reasoning steps. Atom-Searcher~\citep{deng2025atom0searcher0} further decomposes trajectories into fine-grained “atomic thoughts” and trains a reasoning reward model to provide step-level rewards, combining them with outcome rewards in a curriculum to stabilize RL. InfoFlow~\citep{luo2025infoflow0} addresses reward sparsity through reward-density optimization, employing sub-goal scaffolding, expert-guided hints, and dual-agent summarization to inject denser intermediate guidance into RLVR. Our approach differs in both mechanism and scope. Instead of shaping RL rewards, SRR-Judge directly scores and refines each thought–action step. This supervision generalizes beyond controlled RL setups and improves both inference-time deep search and post-training alignment, addressing complex real-world information-seeking tasks rather than multi-hop RAG over fixed corpora.

\subsection{Distilling Evaluation Abilities}

There are abundant works on tailoring LLMs for different types of evaluation tasks~\citep{li-etal-2025-generation}. A central strategy for building efficient and compact judges is through distillation. For example,~\citet{zhu2025judgelm} introduce JudgeLM, which fine-tunes smaller language models using supervision signals from stronger models such as GPT-4o, allowing them to evaluate open-ended generation with high agreement to both teacher and human judgments. Similarly,~\citet{zhang2025distill} use majority-voted LLM judgments to generate reward labels and train compact reward models from these signals. Preference-Aligned Distillation (PAD)~\citep{gu-etal-2025-capturing} further treats the teacher LLM as an implicit reward function and aligns a smaller model to its preference distribution. Additionally, Crowd Comparative Reasoning~\citep{zhang2025crowd} shows that richer CoT judgments from a strong LLM teacher can be leveraged to obtain higher-quality, more robust distilled judges.

Unlike these prior efforts, which primarily distill evaluation capabilities for open-ended text generation, our work targets the agentic deep-search setting. In contrast to judges designed to score static model outputs, SRR-Judge evaluates and refines step-level thought–action pairs within long-horizon search trajectories, offering an scalable way to improve search-integrated reasoning through both inference and post-training alignment.

\section{Conclusion}

We present SRR-Judge, a step-level, model-based evaluator for search-integrated reasoning in long-horizon trajectories. By directly assessing intermediate reasoning and search actions, SRR-Judge provides informative supervision on whether each step meaningfully advances toward a solution. Building on this capability, we introduce a rate-and-refine workflow that is model-agnostic and applicable both for inference-time steering of agent search behavior and for alignment data annotation. When coupled with an iterative rejection-sampling fine-tuning procedure, the SRR-enabled workflow produces high-quality search-alignment data that substantially improves the search-integrated reasoning performance of the base model.

\section*{Limitation}

Although SRR-Judge is effective at evaluating and refining step-level search-integrated reasoning for models without extensive post-training on agentic trajectories, it exhibits limitations when applied to substantially stronger agentic foundation models. As shown in Table~\ref{tab:limitation-comparison}, using SRR-Judge to refine DeepSeek-V3.1 actually reduces performance, indicating that the judge cannot yet supervise a model that is much stronger than itself. This limitation underscores the difficulty of weak-to-strong supervision: current reward models struggle to evaluate or correct the behavior of highly capable agents. Further advances will be needed to make SRR-Judge robust enough to handle such scenarios.

\begin{table}[h!]
\centering
\small
\resizebox{\linewidth}{!}{
\begin{tabular}{lcc}
\toprule
& \textbf{Without SRR} & \textbf{With SRR} \\
\midrule
BrowseComp & 20.7 $\pm$ 0.8 & 17.0 $\pm$ 1.6  \\
BrowseComp-ZH & 50.5 $\pm$ 1.3  & 44.5 $\pm$ 3.6 \\
Xbench-DeepSearch & 67.7 $\pm$ 4.5 & 62.0 $\pm$ 1.4 \\
\midrule
\end{tabular}
}
\caption{Performance degradation when using SRR-Judge to refine outputs of the much stronger DeepSeek-V3.1 model.}
\label{tab:limitation-comparison}
\end{table}

\appendix

\section{Key Prompt Templates}
\label{sec:prompt-template}

Table~\ref{tab:SRR-judge-instruction-prompt} presents the instruction prompt for SRR-Judge to evaluate step-level search-integrated reasoning from four evaluation aspects, including reasoning clarify and conciseness, logical structure, search query appropriateness, and search coverage and reasoning improvement potential.

\begin{table*}[!ht]
\centering
\small
\begin{tabularx}{\textwidth}{|l|X|}
\hline
\textbf{User Query} & \texttt{\{user\_query\}} \\
\hline
\textbf{Previous Reasoning Trace} & \texttt{\{previous\_reasoning\_trace\}} \\
\hline
\textbf{Current Reasoning Trace} & \texttt{\{reasoning\_trace\}} \\
\hline
\textbf{Instruction} & 
\begin{minipage}[t]{0.95\linewidth}
You are now at search iteration \{iteration\}. Please note that:
\begin{itemize}
    \item In \textbf{Previous Reasoning Trace}, the content wrapped with \texttt{<|begin\_search\_query|><|end\_search\_query|>} indicates the queries used to call the search tool; the content wrapped with \texttt{<|begin\_search\_result|><|end\_search\_result|>} indicates the returned search results.
    \item In \textbf{Current Reasoning Trace}, the content enclosed by the \texttt{<|begin\_search\_query|><|end\_search\_query|>} tags indicates either the model’s decision to proceed with the next search plan or the model’s final result wrapped in a box.
\end{itemize}
Given the \textbf{User Query} and \textbf{Previous Reasoning Trace}, score and analyze \textbf{Current Reasoning Trace} based on :
\end{minipage} \\
\hline
\textbf{Evaluation Criteria} & 
\begin{minipage}[t]{0.95\linewidth}
\begin{itemize}
    \item \textbf{Clarity \& Conciseness:} Is the reasoning straightforward, free of redundancy, and easy to understand?
    \item \textbf{Logical Structure:} Are the steps coherent, well-structured, and supportive of the task objective?
    \item \textbf{Query Appropriateness:} Do the search queries accurately reflect the intent, use proper keywords, and retrieve highly relevant information? If the returned content is the final answer, determine whether the final answer is correct and faithful to the information obtained from the previous search results.
    \item \textbf{Coverage \& Improvement Potential:} Does it cover the main points, and is there room for improvement (e.g., adding sources or adjusting strategy)?
\end{itemize}
\end{minipage} \\
\hline
\textbf{Scoring Criteria (1–5)} & 
\begin{minipage}[t]{0.95\linewidth}
\begin{itemize}
    \item \textbf{5 (Excellent):} Clear, well-structured, precise queries, fully covers intent, minimal room for improvement.
    \item \textbf{4 (Good):} Overall strong, with only minor areas for optimization.
    \item \textbf{3 (Fair):} Adequate but contains noticeable redundancy, weaknesses, or gaps.
    \item \textbf{2 (Poor):} Hard to follow, weak logic, queries poorly aligned.
    \item \textbf{1 (Very Poor):} Strongly misaligned with intent, incoherent, little to no value.
\end{itemize}
\end{minipage} \\
\hline
\textbf{Output Format} & 
\begin{minipage}[t]{0.95\linewidth}

Return the result in the following JSON format:

\texttt{X = Integer Score (1–5)}.  

\texttt{Y = Explanation of the score, covering logic, clarity, and query appropriateness.}  

\texttt{Z = An improved version of Current Reasoning Trace (Keep the same first-person reasoning style as Current Reasoning Trace, avoid repeating Previous Reasoning Trace, and do not output any extra words).}  

\texttt{SQ = a new search query. If no further search is needed, output the final result inside \text{\\boxed\{\}}.}

\texttt{{\{"rating": "X", "explanation": "Y", "thought": "Z", "search\_query": "SQ"\}}}
\end{minipage} \\
\hline
\end{tabularx}
\caption{SRR-Judge Instruction Prompt}
\label{tab:SRR-judge-instruction-prompt}
\end{table*}

\section{Additional Qualitative Analysis of SRR-Judge}
\label{sec:qa-srr-additional-example}

Figures~\ref{fig:appendix-srr-artist-step1}, \ref{fig:appendix-srr-artist-step10}, and \ref{fig:appendix-srr-artist-final} together illustrate how SRR-Judge provides consistent process-level supervision across a complete long-horizon search trajectory. At the initial stage, SRR-Judge identifies weak constraint prioritization and encourages the agent to focus on the most distinctive clues rather than speculative enumeration. As the search progresses and the agent becomes stuck in repeated, unproductive poet-centric queries, SRR-Judge intervenes again at Step 10, detecting long-horizon drift and redirecting the search toward a more discriminative institutional clue—the art school founded in the 1980s. This pivot enables rapid identification of the correct artist. Finally, once all constraints are systematically verified, SRR-Judge assigns a high score, recognizing the coherence and completeness of the reasoning and requiring only minimal confirmation to finalize the answer.

\begin{figure}[!t]
\footnotesize
\fbox{
\begin{minipage}{0.96\linewidth}

\textbf{Question.} \\
The artist’s father was a Protestant minister. They had a brief marriage to a colloquial poet who studied engineering and was born in September between 1938--1941. Later, they married a fashion designer. In the 1980s, they established a school for the art they were known for in a village where that art had not previously existed. What is the artist’s name?

\medskip
\textbf{Answer.} \\
Evelyne Porret

\medskip
\textbf{History (Steps 1--9; summarized).} \\
\textbf{Step 1 (Search).} \texttt{colloquial poet born September 1938 engineering degree} \\
\textbf{Obs. 1.} No direct match; partial leads (e.g., Jack Adair; Vasudev Nirmal). \\
\textbf{Step 2 (Search).} \texttt{Vasudev Nirmal date of birth} \\
\textbf{Obs. 2.} Most sources indicate June 2, 1936 (outside target range). \\
\textbf{Steps 3--6.} Repeated poet-centric searches with tighter constraints; no match found; keyword drift (e.g., ``September 1939'' $\rightarrow$ W.H. Auden). \\
\textbf{Step 7 (Search).} \texttt{artist Protestant minister father married fashion designer established art school 1980s village} \\
\textbf{Obs. 7.} No direct match; only partial/irrelevant hits. \\
\textbf{Steps 8--9.} Additional poet searches; still no match.

\medskip
\textbf{Original Reasoning and Search Action (Step 10).} \\
``Given the difficulty in pinpointing the poet, perhaps the key lies in the art school established in the 1980s in a village where the art hadn't existed before. [...]
\textcolor{red}{This is getting too speculative. Let's return to the original clues.} [...]
\textcolor{red}{The key might be to find the poet first.} [...]
\textcolor{red}{Possible candidates:} [...]
\textcolor{red}{None of these poets fit the exact criteria. Perhaps the poet is less well-known.} [...]
\textcolor{red}{Let me search for poets born September 1938--1941.}'' \\
\textit{Search Query:} \texttt{poets born September 1938 to 1941}

\medskip
\textbf{SRR-Judge Evaluation.} \\
\textit{Rating: 2/5.} \\
``The current reasoning trace \textcolor{blue}{lacks clarity and conciseness}, as it meanders through speculative lists without a focused strategy. The logical structure is weak and does not build coherently on previous searches. The search query \textcolor{blue}{omits critical criteria like the engineering degree and colloquial style}, reducing relevance.''

\medskip
\textbf{SRR-Judge Refined Thought and Action.} \\
``\textcolor{blue}{I need to find the artist by focusing on the art school establishment clue, which may be more distinctive. The artist founded a school for their art in the 1980s in a village where that art was previously absent. I'll search for such schools and identify the founder, then verify the remaining clues (Protestant minister father; brief marriage to a colloquial engineer-poet born in September 1938--1941; later marriage to a fashion designer).}'' \\
\textit{Search Query:} \texttt{pottery school founded 1980s village}

\end{minipage}
}
\caption{Artist Example Step 10 - showing SRR-Judge intervening after prolonged search failure. Highlighted text (red) marks a degeneration into speculative enumeration and an under-specified query; SRR-Judge redirects the agent to pivot on a more searchable institutional clue (the art school) and then use the remaining biographical constraints for verification.}
\label{fig:appendix-srr-artist-step10}
\end{figure}

\begin{figure}[!t]
\footnotesize
\fbox{
\begin{minipage}{0.96\linewidth}

\textbf{Question.} \\
The artist’s father was a Protestant minister. They had a brief marriage to a colloquial poet who studied engineering and was born in September between 1938--1941. Later, they married a fashion designer. In the 1980s, they established a school for the art they were known for in a village where that art had not previously existed. What is the artist’s name?

\medskip
\textbf{Answer.} \\
Evelyne Porret

\medskip
\textbf{History (Steps 10--12; summarized).} \\
\textbf{Step 10 (Search).} \texttt{pottery school founded 1980s village} \\
\textbf{Obs. 10.} Identifies \textbf{Evelyne Porret} as founder of a pottery school in the village of Tunis, Egypt, in the 1980s. \\
\textbf{Step 11 (Search).} \texttt{Evelyne Porret biography husband} \\
\textbf{Obs. 11.} Confirms: father was a Protestant pastor; first marriage to colloquial poet Sayed Hegab; later marriage to a fashion designer. \\
\textbf{Step 12 (Search).} \texttt{Sayed Hegab birth date education} \\
\textbf{Obs. 12.} Confirms Hegab was born September 23, 1940, and studied engineering.

\medskip
\textbf{Original Reasoning and Search Action (Step 13).} \\
``\textcolor{red}{All clues are now satisfied:} [...] \\
\textcolor{red}{Thus, the artist's name is Evelyne Porret.} [...]
\textcolor{red}{But I should confirm the commonly used name.}'' \\
\textit{Search Query:} \texttt{Evelyne Porret full name}

\medskip
\textbf{SRR-Judge Evaluation.} \\
\textit{Rating: 5/5.} \\
``The current reasoning trace is \textcolor{blue}{clear and concise}, summarizing the successful verification of all clues without redundancy. The logical structure is coherent, building on previous searches to confirm Evelyne Porret as the artist. The query is \textcolor{blue}{highly appropriate} and coverage is complete.''

\medskip
\textbf{SRR-Judge Refined Thought and Action.} \\
``\textcolor{blue}{Evelyne Porret is the artist, as all clues align. I will confirm her full name to finalize the answer.}'' \\
\textit{Search Query:} \texttt{Evelyne Porret}

\end{minipage}
}
\caption{Artist Example Final Step - showing SRR-Judge recognizing successful constraint satisfaction at the final stage and assigning a high score. Highlighted text (red) indicates concise synthesis and minimal confirmation needed to finalize the answer.}
\label{fig:appendix-srr-artist-final}
\end{figure}

Another example is presented in figures~\ref{fig:qualitative-srr-racer},~\ref{fig:appendix-srr-step3}, and~\ref{fig:appendix-srr-racer-final}. At the start (Step~1), the agent correctly prioritizes the rare high-school founding-date constraint, but the initial query yields no actionable evidence, necessitating a change in the search strategy. After Step~2 produces a bounded candidate set (AAA champions from 1935–1940), SRR-Judge steers the agent toward a disciplined, step-by-step verification process (Step~3), systematically checking each candidate against remaining criteria and preventing premature speculation or candidate hopping. As the trajectory progresses and multiple distinctive clues (e.g., August birth, cremation, childhood anecdote) fail to resolve the search, SRR-Judge continues to reward structured reasoning and penalize diffuse exploration. In the final stage, when the search budget is exhausted, SRR-Judge assigns a high score to the search agents' synthesis of accumulated evidence and a principled pivot back to the rare high-school constraint, refining the query toward race-driver alumni rather than unstructured guessing. Finally, the agent makes a best-effort guess grounded in verified facts. 

These examples demonstrate how SRR-Judge supports both mid-trajectory correction and late-stage re-planning, promoting constraint-driven search, graceful degradation under tool limits, and robust long-horizon reasoning.

\begin{figure}[!t]
\small
\fbox{
\begin{minipage}{0.96\linewidth}

\textbf{Question.} \\
There’s an American racer who was born in August and was cremated when he died. He won his first title between 1935--1940, earned a pole position in the 1950s, attended a high school founded between 1800--1810, and drove his father’s car at age 5. What was his age when he participated in his first race?

\medskip
\textbf{Answer.} \\
17

\medskip
\textbf{Original Reasoning and Search Action (Step 1).} \\
``Okay, let's try to figure this out step by step. The question is asking about an American racer who meets several specific criteria. [...]

First, the racer was born in August and was cremated when he died. [...] He won his first title between 1935 and 1940. [...] His high school was founded between 1800--1810. [...]

\textcolor{red}{Now, putting together these clues. Let me think of famous racers from that period. A.J. Foyt comes to mind... Mario Andretti... Bill Vukovich... Rex Mays... Wilbur Shaw...}

\textcolor{red}{Let me do a search for American racers who fit the criteria.}

Searching for ``American race car driver born August, cremated, first championship 1935--1940''.'' \\
\textit{Search Query:} \texttt{American race car driver born August, cremated, first championship 1935--1940}

\medskip
\textbf{SRR-Judge Evaluation.} \\
\textit{Rating: 3/5.} \\
``The reasoning trace is fairly clear in breaking down the clues but contains \textcolor{blue}{speculative jumps between racer names} without a structured approach. The logical structure is weak as it \textcolor{blue}{does not prioritize or systematically use the clues}. The search query is partially appropriate but \textcolor{blue}{omits key criteria such as the high school foundation date and pole position in the 1950s}.''

\medskip
\textbf{SRR-Judge Refined Thought and Action.} \\
``\textcolor{blue}{The high school founded between 1800--1810 is a precise and discriminative clue. I should search for this first, then cross-reference with other constraints.}'' \\
\textit{Search Query:} \texttt{American race car driver high school founded 1800--1810}

\end{minipage}
}
\caption{American Racer Example Step 1 - showing how SRR-Judge identifies speculative candidate enumeration and poor constraint prioritization (highlighted in red), and refines the search strategy toward more discriminative evidence.}
\label{fig:qualitative-srr-racer}
\end{figure}

\begin{figure}[!t]
\small
\fbox{
\begin{minipage}{0.96\linewidth}

\textbf{Question.} \\
There’s an American racer who was born in August and was cremated when he died. He won his first title between 1935--1940, earned a pole position in the 1950s, attended a high school founded between 1800--1810, and drove his father’s car at age 5. What was his age when he participated in his first race?

\medskip
\textbf{Answer.} \\
17

\medskip
\textbf{History (Steps 1--2; summarized).} \\
\textbf{Step 1 (Search).} \texttt{American race car driver high school founded 1800--1810} \\
\textbf{Obs. 1.} No relevant results; query returns unrelated local history / documents. \\

\textbf{Step 2 (Search).} \texttt{american race car drivers first championship 1935--1940} \\
\textbf{Obs. 2.} Identifies AAA national champions in-window: \textit{Wilbur Shaw (1937), Floyd Roberts (1938), Jimmy Snyder (1939), Tony Willman (1940)}.

\medskip
\textbf{Original Reasoning and Search Action (Step 3).} \\
``Alright, so possible names include Wilbur Shaw, Floyd Roberts, Jimmy Snyder, and Tony Willman. Now, let's check each of these drivers for the other criteria. [...]

\textcolor{red}{Wait, perhaps someone else? Another approach: drivers who had a pole position in the 1950s.} [...] 
\textcolor{red}{Maybe Bill Vukovich?} [...] 
\textcolor{red}{Phil Hill?} [...] 
\textcolor{red}{Paul Russo? Not sure.}

\textcolor{red}{Alternatively, looking for drivers cremated.} For example, Rex Mays was cremated, but he was born in July. [...]

\textcolor{red}{Maybe I'm missing someone. Let's try combining ``racer cremated august birth'' and similar terms.}'' \\
\textit{Search Query:} \texttt{american race car driver cremated august born}

\medskip
\textbf{SRR-Judge Evaluation.} \\
\textit{Rating: 2/5.} \\
``The reasoning trace is poorly structured and lacks clarity, as it \textcolor{blue}{jumps between drivers without a systematic approach} and \textcolor{blue}{dismisses candidates based on incomplete checks}. It also \textcolor{blue}{fails to address the high school clue}, a key filter. The proposed search query is \textcolor{blue}{vague and misaligned with the intent}, ignoring critical criteria such as pole position in the 1950s.''

\medskip
\textbf{SRR-Judge Refined Thought and Action.} \\
``\textcolor{blue}{I need to systematically verify each AAA national champion from 1935--1940 for all criteria. Jimmy Snyder was born in August 1918 and won his first title in 1939, matching two key points. I should search for his biography to confirm the remaining details.}'' \\
\textit{Search Query:} \texttt{Jimmy Snyder biography cremated pole position 1950s high school founded date}

\end{minipage}
}
\caption{American Racer Example Step 3 - SRR-Judge intervening at a late stage when search degenerates into speculative enumeration. Highlighted text (red) marks unstructured reasoning and the judge’s corrective guidance toward systematic verification.}
\label{fig:appendix-srr-step3}
\end{figure}

\begin{figure}[!t]
\footnotesize
\fbox{
\begin{minipage}{0.96\linewidth}

\textbf{Question.} \\
There’s an American racer who was born in August and was cremated when he died. He won his first title between 1935--1940, earned a pole position in the 1950s, attended a high school founded between 1800--1810, and drove his father’s car at age 5. What was his age when he participated in his first race?

\medskip
\textbf{Answer.} \\
17

\medskip
\textbf{History (selected; summarized).} \\
\textbf{Step 1 (Search).} \texttt{American race car driver high school founded 1800-1810} \\
\textbf{Obs. 1.} No relevant results. \\
\textbf{Step 2 (Search).} \texttt{american race car drivers first championship 1935-1940} \\
\textbf{Obs. 2.} AAA champions shortlist: Wilbur Shaw (1937), Floyd Roberts (1938), Jimmy Snyder (1939), Tony Willman (1940). \\
\textbf{Steps 3--6.} Systematic verification rules out AAA champions (e.g., died before 1950s; mismatch on August/cremation/pole). \\
\textbf{Steps 7--19.} Multiple pivots (August+title search; childhood anecdote search; candidate checks like Agabashian), but high-school founding constraint remains unresolved.

\medskip
\textbf{Final Search Step Reasoning.} \\
``\textcolor{red}{None of the drivers found so far meet all criteria. Perhaps the answer requires integrating all known clues. Let me recap:} [...]
\textcolor{red}{Given the challenges, perhaps the high school is the key. High schools founded between 1800--1810 in the US are rare.} \\
\textcolor{red}{Let me search for a list of high schools in that time frame and see if any notable alumni include racers fitting the other criteria.}'' \\
\textit{Search Query:} \texttt{List of high schools in the United States founded between 1800-1810}

\medskip
\textbf{SRR-Judge Evaluation.} \\
\textit{Rating: 4/5.} \\
``The reasoning is \textcolor{blue}{clear and concise}, effectively summarizing the dead ends and pivoting to a new strategy focused on the high school clue. The logical structure is coherent. The query is appropriate for gathering a list of schools, but it could be improved by \textcolor{blue}{incorporating `race car driver alumni'} to retrieve more relevant results.''

\medskip
\textbf{SRR-Judge Refined Thought and Action.} \\
``\textcolor{blue}{I'll search for a list of high schools founded between 1800--1810 to identify potential alumni who are race car drivers. Then, I can cross-reference with the other criteria.}'' \\
\textit{Search Query:} \texttt{American race car driver alumni of high schools founded 1800-1810}

\medskip
\textbf{Final Answer Stage (Search Budget Exhausted).} \\
\textbf{Step 20 Obs.} ``\textcolor{red}{You have reached the search limit. You are not allowed to search.}'' \\

\textbf{Step 21 (Answer).} ``\textcolor{red}{Since I can't proceed further with additional searches, I need to rely on existing knowledge.} [...] \\
\textcolor{red}{For Fred Agabashian,} [...] \textcolor{red}{he started racing jalopies at 17 in 1931} [...] \\
\textcolor{red}{Thus, the most plausible answer, albeit with incomplete data, is 17.}''

\end{minipage}
}
\caption{American Racer Final Step - After multiple dead ends, SRR-Judge rewards a structured recap and a pivot to a more discriminative constraint (rare high-school founding dates), while suggesting a more targeted query. Though the search budget is exhausted, the agent provides the most plausible answer.}
\label{fig:appendix-srr-racer-final}
\end{figure}

\end{document}